\documentclass[graybox]{svmult}

\usepackage[numbers,sort&compress]{natbib}
\usepackage{amsmath}
\usepackage{bm}
\usepackage{amsfonts}
\usepackage{float}
\usepackage{latexsym}

\usepackage{caption} 
\usepackage{subcaption} 
\usepackage{wrapfig} 

\renewcommand{\vec}[1]{\mathbf{#1}}
\newcommand{\transdens}{p\left(\vec{x}_{t+1} \,\vert\, \vec{x}_t, \vec{u}_t\right)}
\newcommand{\obsdens}{p\left(\vec{y}_{t} \,\vert\, \vec{x}_t, \vec{u}_t\right)}

\usepackage{type1cm}        
\usepackage{makeidx}         
\usepackage{graphicx}        
\usepackage{multicol}        
\usepackage[bottom]{footmisc}

\usepackage{newtxtext}       %
\usepackage{newtxmath}       

\makeindex             

\begin{document}

\title*{Physics-informed machine learning for Structural Health Monitoring}
\author{E. J. Cross, S. J. Gibson, M. R. Jones, D. J. Pitchforth, S. 
Zhang and T.J. Rogers}
\institute{E. J. Cross, S. J. Gibson, M. R. Jones, D. J. Pitchforth, S. 
Zhang, T.J. Rogers \at Dynamics Research Group, Department of Mechanical Engineering, University of Sheffield, Mappin St, Sheffield, S1 3JD UK. \email{e.j.cross@sheffield.ac.uk}}
%
%
\maketitle

\abstract*{The use of machine learning in Structural Health Monitoring is 
becoming more 
common, as many of the inherent tasks (such as regression and classification) 
in developing condition-based assessment fall naturally into its remit. This 
chapter introduces the concept of physics-informed machine learning, 
where 
one adapts ML algorithms to account for the physical insight an engineer will 
often have of the structure they are attempting to model or assess. The 
chapter will demonstrate how grey-box models, that combine simple 
physics-based models with data-driven ones, can improve predictive capability 
in an SHM setting. A particular strength of the approach demonstrated here 
is the capacity of the models to generalise, with enhanced predictive 
capability in different regimes. This is a key issue when life-time 
assessment is a requirement, or when monitoring data do not span the 
operational conditions a structure will undergo. \newline
The chapter will provide an overview of 
physics-informed 
ML, 
introducing a 
number of new approaches for grey-box modelling in a Bayesian setting. The 
main ML tool discussed will be Gaussian process regression, we will 
demonstrate how physical assumptions/models can be incorporated through 
constraints, through the mean function and kernel design, and finally in a state-space setting. A 
range of SHM applications will be demonstrated, from loads monitoring tasks 
for off-shore and aerospace structures, through to performance 
monitoring for 
long-span bridges. }

\abstract{The use of machine learning in Structural Health Monitoring is 
becoming more 
common, as many of the inherent tasks (such as regression and classification) 
in developing condition-based assessment fall naturally into its remit. This 
chapter introduces the concept of physics-informed machine learning, 
where 
one adapts ML algorithms to account for the physical insight an engineer will 
often have of the structure they are attempting to model or assess. The 
chapter will demonstrate how grey-box models, that combine simple 
physics-based models with data-driven ones, can improve predictive capability 
in an SHM setting. A particular strength of the approach demonstrated here 
is the capacity of the models to generalise, with enhanced predictive 
capability in different regimes. This is a key issue when life-time 
assessment is a requirement, or when monitoring data do not span the 
operational conditions a structure will undergo. \newline
The chapter will provide an overview of 
physics-informed 
ML, 
introducing a 
number of new approaches for grey-box modelling in a Bayesian setting. The 
main ML tool discussed will be Gaussian process regression, we will 
demonstrate how physical assumptions/models can be incorporated through 
constraints, through the mean function and kernel design, and finally in a state-space setting. A 
range of SHM applications will be demonstrated, from loads monitoring tasks 
for off-shore and aerospace structures, through to performance 
monitoring for 
long-span bridges. }

\section{Introduction}

As performance and monitoring data from our structures become more abundant, 
it is natural for researchers to turn to methods from the machine learning 
community to help with analysis and construction of diagnostic/prognostic 
algorithms. 
Indeed, within the SHM research field, use of neural networks, 
support vector machines and Gaussian processes for regression and 
classification problems has become common place 
\cite{farrar2010introduction}. These methods bring the 
opportunity to learn complex relationships directly from 
data, without a requirement of in-depth knowledge of the system. As an 
example from 
the authors' own work, in \cite{holmes2016prediction} we employed a Gaussian 
process (GP) regression to predict strain on a landing gear from measured 
accelerations across the aircraft. Use of a suitably trained GP circumvents 
the need to build complex physics-based models of the gear for fatigue life 
calculations. This kind of model is often referred to as a `black-box' model 
to reflect the fact the data drives the structure of the model rather than 
knowledge of the physics at work. 

At the other end of the spectrum the term 
`white-box' model can be used to describe a model purely constructed from 
knowledge of physics, (e.g differential equations and finite element 
models). Physics-based modelling and updating were common early themes in   
the structural health monitoring research field \cite{sohn2003review}. 
However, for large or critical engineering structures that operate in (often 
extreme) 
dynamic environments, such as wind turbines, aircraft, gas turbines,
etc, predictive modelling from a white-box perspective presents 
particularly difficult challenges. Loading is often unknown and unmeasured, 
and dynamic behaviour during operation needs to be fully captured by a 
computational model, but is sensitive to small changes in (or 
disturbances to) the structure. Validation and updating of large complex 
models bring their own 
challenges and remain active research areas 
\cite{simoen2015dealing,gardner2018novel,gardner2019unifying,gardner2019bayesian}.

\begin{figure}[h]
\centering
\begin{subfigure}{0.48\linewidth}
\includegraphics[width=\linewidth]{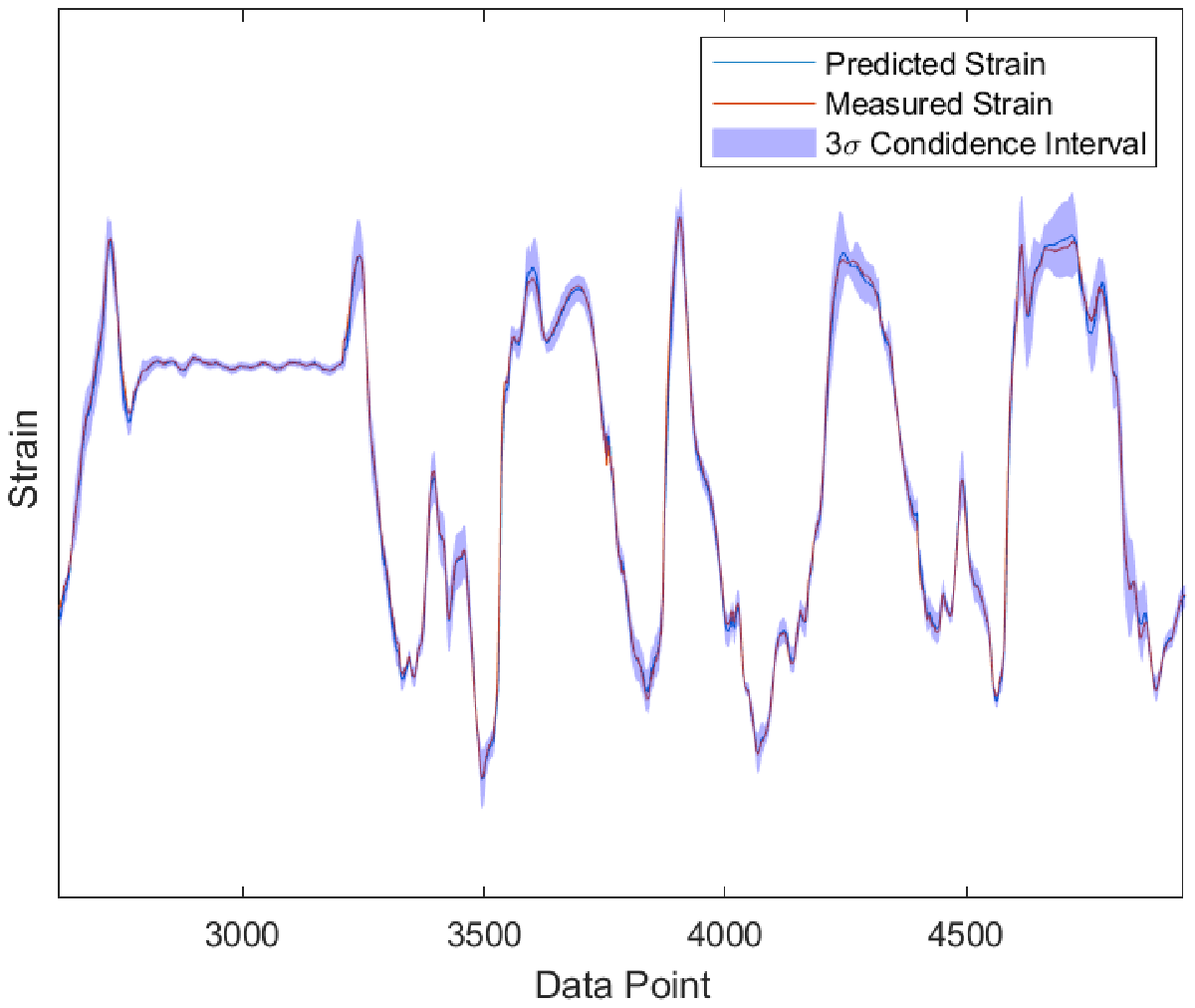}
\caption{Prediction for normal flight conditions}
\label{fig:tucano_good}
\end{subfigure}
\begin{subfigure}{0.48\linewidth}
\includegraphics[width=\linewidth]{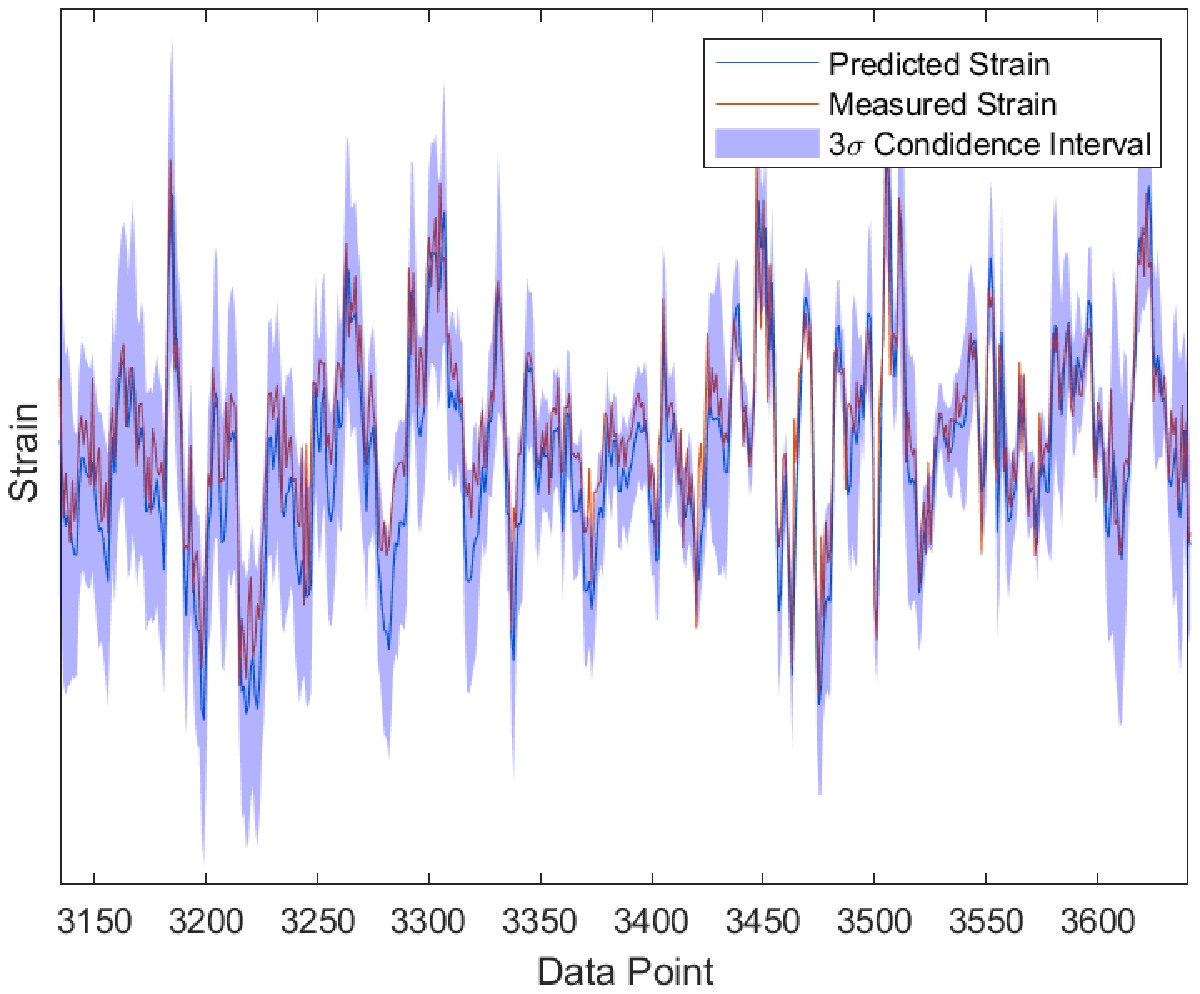}
\caption{Prediction for low altitude flight}
\label{fig:tucano_low}
\end{subfigure}
\caption{Example of difficulty predicting behaviour outside of normal 
operation conditions - strain prediction on an aircraft wing during 
manoeuvres - see \cite{fuentes2014aircraft,gibson2020data} for more 
details.}
\label{fig:tucano}
\end{figure}

Due to the availability of monitoring data, the inherent challenges of the 
physics-based approach and the promise of machine learning methods, it is 
fair to say that the data-driven approach to SHM has become dominant in the 
research field. A 
significant issue with the use of any machine 
learning method in an engineering application, however, is the availability 
of suitable data with which to train the algorithm. As the model learns from 
the data, it is only able to accurately predict behaviour present in the data 
on which 
it was trained. As an example, Figure \ref{fig:tucano} shows a black-box 
model trained to 
predict the bending strain on an aircraft wing during different manoeuvrers 
to inform an in-service fatigue assessment. This data set comprises of 84 
flights, five of which are used for model training. The trained model is 
able to generalise well with a very low prediction error for the majority 
of the flights  - 
Figure \ref{fig:tucano_good} shows a typical strain prediction for a flight 
not included in the 
training set (normalised mean-squared error\footnote{\begin{equation}
\label{eqn:MSE}
\textrm{nMSE} = \tfrac{100}{n\sigma_y^2}\sum{(y_i - f_i)^2}
\end{equation}
where $y_{i}$ and $f_{i}$ are the measurements and predictions 
respectively, $i=1\dots n$.} (nMSE)= 0.29\% across the whole flight). 
However, for the flight 
shown in Figure \ref{fig:tucano_low}, 
the model is unable 
to predict the strain as accurately (nMSE 4.20\% across the whole flight) - 
this flight was atypical in terms of 
operating conditions - it was a low altitude sortie over ground, 
characterised by the turbulent response one can see in the figure. These 
conditions are different from those included in the training set and the 
model is 
unable to generalise and predict the strain as well in this case.

In general, but especially because of the inherent flexibility in many of the 
machine learning models commonly used, 
extrapolation should not be attempted in this setting. For an SHM 
application, 
this will 
generally mean that training data are required from all possible operating 
conditions that the structure will see. For many applications this is 
currently infeasible, although as data collection becomes more commonplace, 
the situation will improve somewhat. Where a supervised approach is needed, 
this problem is exacerbated by the general lack of access to data from 
structures in a 
damaged state which remains a large barrier to effective diagnosis and 
prognosis \cite{barthorpe2010model}.

Currently a programme of work by the authors is pursuing a physics-informed 
machine learning approach to attempt to address some of these issues in a 
structural dynamics setting. The aim is to bring together the flexibility and 
power 
of state-of-the-art machine learning techniques with more structured and 
insightful physics-based models derived from domain expertise. This reflects 
a natural wish that any inferences over our structures will be informed by 
both our engineering knowledge and relevant monitoring data available. 

The potential 
means of combining physics-based models and data-driven algorithms are many, 
ranging from employing ML methods for parameter estimation 
\cite{farrar2007nonlinear,kerschen2006past}, to using 
them as 
surrogates or emulators 
\cite{oakley2002bayesian,leser2017probabilistic,Beucler2019}. Of 
interest here are methods where the 
explanatory power of a model is shared between physics-based and data-driven 
components. We will often refer to these approaches as 
`grey-box' models (a combination of white and black-box components), but the 
term `hybrid modelling' is equally applicable. The philosophy followed in our 
work is to embed fundamental 
physical insight 
into a machine learning algorithm. In doing so, our aim is that the role of 
the machine learner is one of augmenting the explanatory power of the model 
rather than being employed to correct any potential error or bias in the 
physical foundation. This chapter will explore this idea in a Bayesian 
setting, introducing a number of different approaches and demonstrating their 
usefulness in an SHM setting.

\section{Grey-box models, overview and literature}
\label{sec:lit}

\begin{wrapfigure}{r}{0.39\textwidth}
\centering
\includegraphics[width=\linewidth]{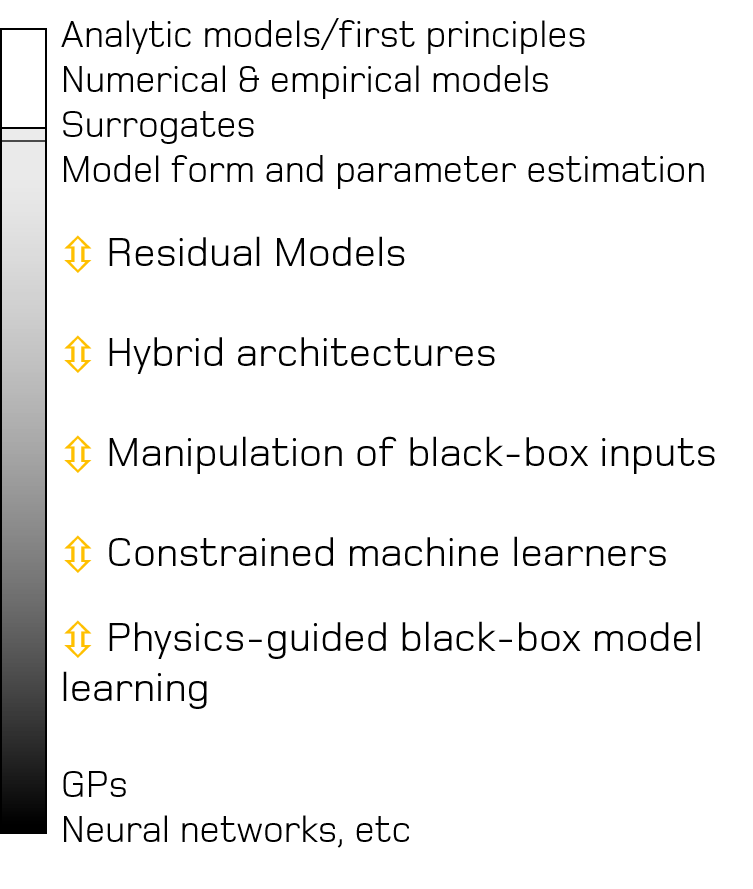}
\caption{Some modelling approaches on the white-black spectrum.}
\label{fig:spectrum}
\end{wrapfigure}
The term `grey-box model' is perhaps most familiar to those from a control engineering background. Sohlberg \cite{sohlberg2008grey,sohlberg2012supervision} provides a useful review and overview of grey-box models in this context\footnote{The use of grey-box models within the control community is undergoing somewhat of a revival there and a good snap-shot of this 
can be gained by looking at the contributions to the most recent Nonlinear 
System Identification Benchmarks Workshop (http://www.nonlinearbenchmark.org)}. Figure \ref{fig:spectrum} attempts to capture and summarise some of the currently available 
modelling approaches relevant for challenges in structural health monitoring 
on the white to black spectrum. Note that the ``degree of greyness" of the models in the middle region will change according to implementation and application. 

At the whiter end of the spectrum are modelling approaches where data are 
used for parameter estimation or model form selection, (with the buoyant 
field of equation discovery fitting in here, see 
\cite{fuentes2021equation,brunton2016discovering}). See also 
\cite{noel2018grey}. \textit{Residual models} are those that use a 
data-driven approach to account for the observed difference between a 
physics-based model and measurements, with general form 
\begin{equation}
y = \!\! \underbrace{f(x)}_{White-box} \!\!\! + \;\: \underbrace{\delta (x) +  \varepsilon}_{Black-box}
\label{eq:rm}
\end{equation}
where $f(x)$ is the output of the physical model, $\delta (x)$ is the model discrepancy and $\varepsilon$ is the process noise (see for example \cite{kennedy2001bayesian,lei2020considering,brynjarsdottir2014learning,worden2007identification}). The discrepancy term is often used to correct a misspecified physical model, giving rise to the term `bias correction'. Residual based approaches have proven effective across a range of SHM tasks including damage detection \cite{BiasCorrectionFELamb} and modal identification \cite{BiasCorrectionFEGP}. Here we are interested in residual modelling in the context of compensation for un-captured/missing behaviours in the physics-based model (discussed further in Section \ref{sec:bayes}). 

The term \textit{hybrid architectures} reflects the wider possibilities for 
combinations of white and black models (which could include the summation 
form of (\ref{eq:rm})). Section \ref{sec:ss} will demonstrate one such 
example of combining data-driven and physics based models in a state-space 
setting. 

The remainder of the spectrum contains models with structures that are 
data-driven/black-box in nature. Sohlberg \cite{sohlberg2008grey}) describes 
\textit{semi-physical modelling} as when features are subject to a nonlinear 
transformation before being used as inputs to a black-box model, we also 
refer to this as \textit{input augmentation}, see 
\cite{rogers2017grey,worden2018evolutionary,fuentes2014aircraft} for more 
examples. We place these examples under the heading of \textit{manipulation 
of black-box inputs}. 

Section \ref{sec:constrain} of this chapter will 
discuss \textit{constraints} for machine learning algorithms - these are 
methods that 
allow one to constrain the predictions of a machine learner so that they 
comply 
with physical assumptions. Excellent examples for Gaussian process regression 
are  
\cite{solin2018modeling, wahlstrom2013modeling,jidling2018probabilistic} and 
will be discussed in greater detail later.

The final grouping of grey-box approaches 
mentioned here are physics-guided black-box learners. These are methods that 
use physical insight to attempt to improve model optimisation and include the 
construction of physics-guided loss functions and the use of physics-guided
initialisation. These will not be discussed further in this chapter but see e.g. \cite{Karpatne2017a,Raissi2019,Karpatne2017,willard2020integrating} for more details.

\subsubsection{Grey-box models for SHM} 
The remainder of the chapter will showcase some of the work of the authors on 
developing physics-informed machine learning approaches for SHM tasks. The 
developments here fall in the domain of residual and hybrid 
models (Sections \ref{sec:bayes} and \ref{sec:ss}), and constrained machine 
learners (Section \ref{sec:constrain}). In 
order to provide an overview, a variety of methods and results are presented, 
however, the implementation details given here are necessarily very brief and 
we refer readers to the 
referenced papers and our webpage\footnote{https://drg-greybox.github.io/} 
for specific details 
and more in-depth analysis.
Reflecting the philosophy discussed in the introduction section, the 
approaches presented generally incorporate simple physics-based models or 
assumptions 
and rely on the machine learner for enhanced explanatory power and 
flexibility (i.e. we are operating towards the blacker end of the scale).  

The machine learning approach used in the 
work shown here will be Gaussian process (GP) regression throughout. GPs have 
been 
shown to be a powerful tool for regression tasks \cite{Rasmussen2006} and are 
becoming common in SHM applications (see for example 
\cite{holmes2016prediction,bull2020probabilistic,wan2018bayesian,
kullaa2011distinguishing,rogers2020application}).  Their 
use here and throughout the work of the authors is due to their 
(semi)non-parametric nature, their ability to function with a small number of 
training points, and most importantly, the Bayesian framework within which 
they 
naturally work. The Gaussian process 
formulation provides a predictive 
\textit{distribution} rather than a single prediction point, allowing 
confidence intervals to be calculated and uncertainty to be propagated 
forward into any following analysis (see \cite{gibson2020data} for example). 
As the use of GPs is now quite common, their fundamental formulation will not 
be introduced here, but the mathematical machinery required is 
briefly summarised in the Appendix - we refer unfamiliar readers to 
\cite{Rasmussen2006}.

In the first examples shown here, the use of priors in the Bayesian framework 
is exploited as an appropriate and intuitive means of incorporating physical 
insight into a machine learning algorithm. In later sections we consider the 
construction of constraints for GPs and, separately, their incorporation into 
a state-space formulation (this latter example relies more heavily on 
physics-based machinery than the other examples).

\section{Be more Bayes}
\label{sec:bayes}

A Bayesian philosophy is one that employs evidence from data to update prior 
beliefs or 
assumptions, and has been widely adopted across disciplines, including SHM. 
However, 
most commonly, uninformative priors are utilised that do not reflect the
knowledge that we have as engineers of the systems we are interested in 
modelling. 

The formulation of a Gaussian process regression requires the selection of a  
mean and covariance function which form the prior process. The process is 
then conditioned with training data to provide a posterior mean and 
covariance as 
the model prediction. In the standard approach, no prior knowledge is assumed; 
a zero mean function is selected alongside a generic covariance function such 
as a squared-exponential or one from the Mat\'{e}rn class which provide a 
flexible process to fit to most data.

In this section we will first employ simple physical models as prior mean 
functions to a GP and show how they may improve the extrapolative capability 
of 
the model. This simple means of incorporating prior knowledge is equivalent to using a GP with a zero mean prior to model the difference between the measured data and the physical model prediction, and hence can be 
classed as a residual approach (see Section \ref{sec:lit}). At the end of 
this section, we will show how some knowledge of a system may be 
used to derive useful covariance functions in a regression setting.

\subsection{Prior mean functions - residual modelling}

\subsubsection{Performance monitoring of a cable-stayed bridge}
\begin{figure}[h]
\centering
\begin{subfigure}[b]{0.45\textwidth}
\centering
\includegraphics[width=1\linewidth]{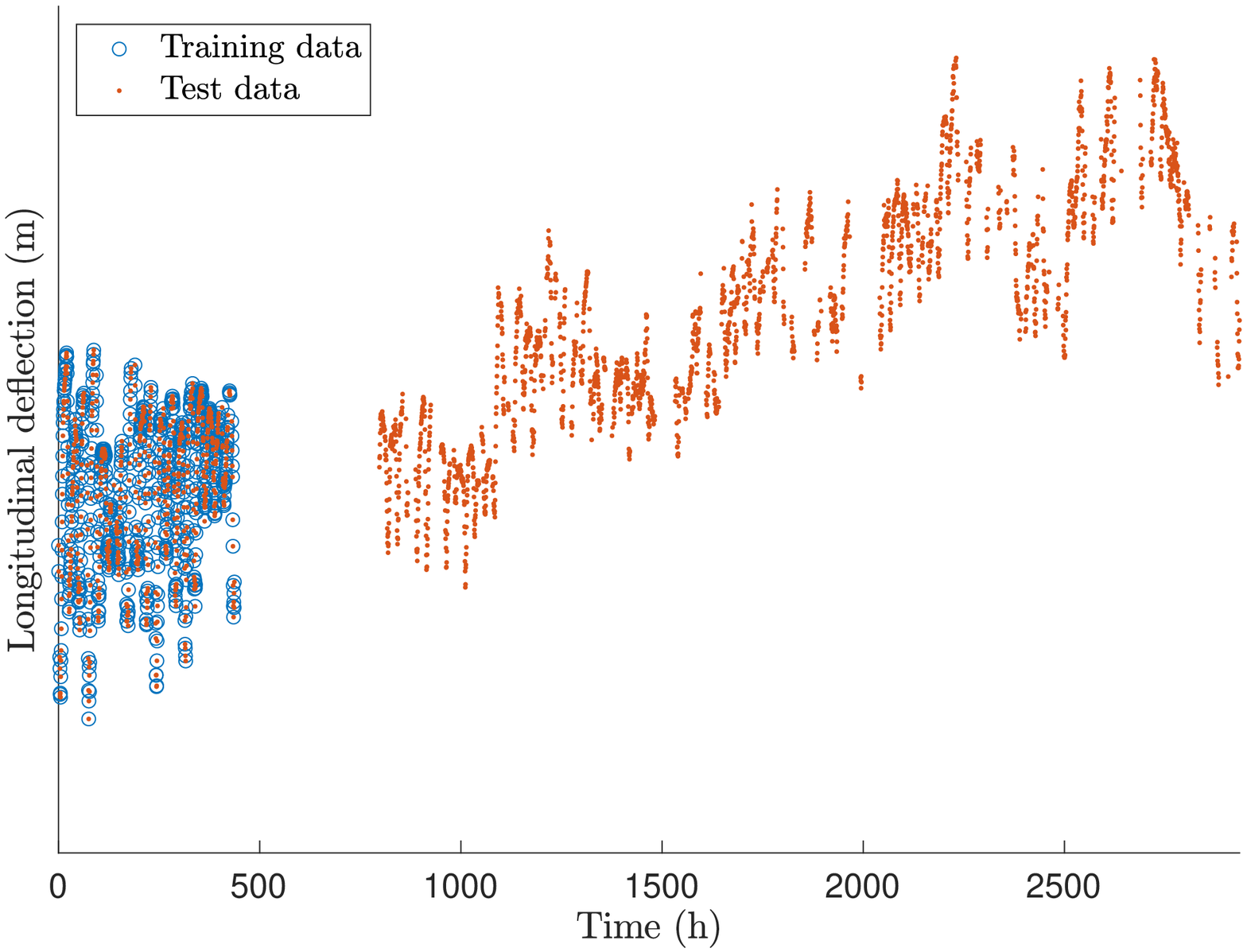}
\caption{}
\label{fig:tamar_data}
\end{subfigure}
\begin{subfigure}[b]{0.45\textwidth}
\centering
\includegraphics[width=1\linewidth]{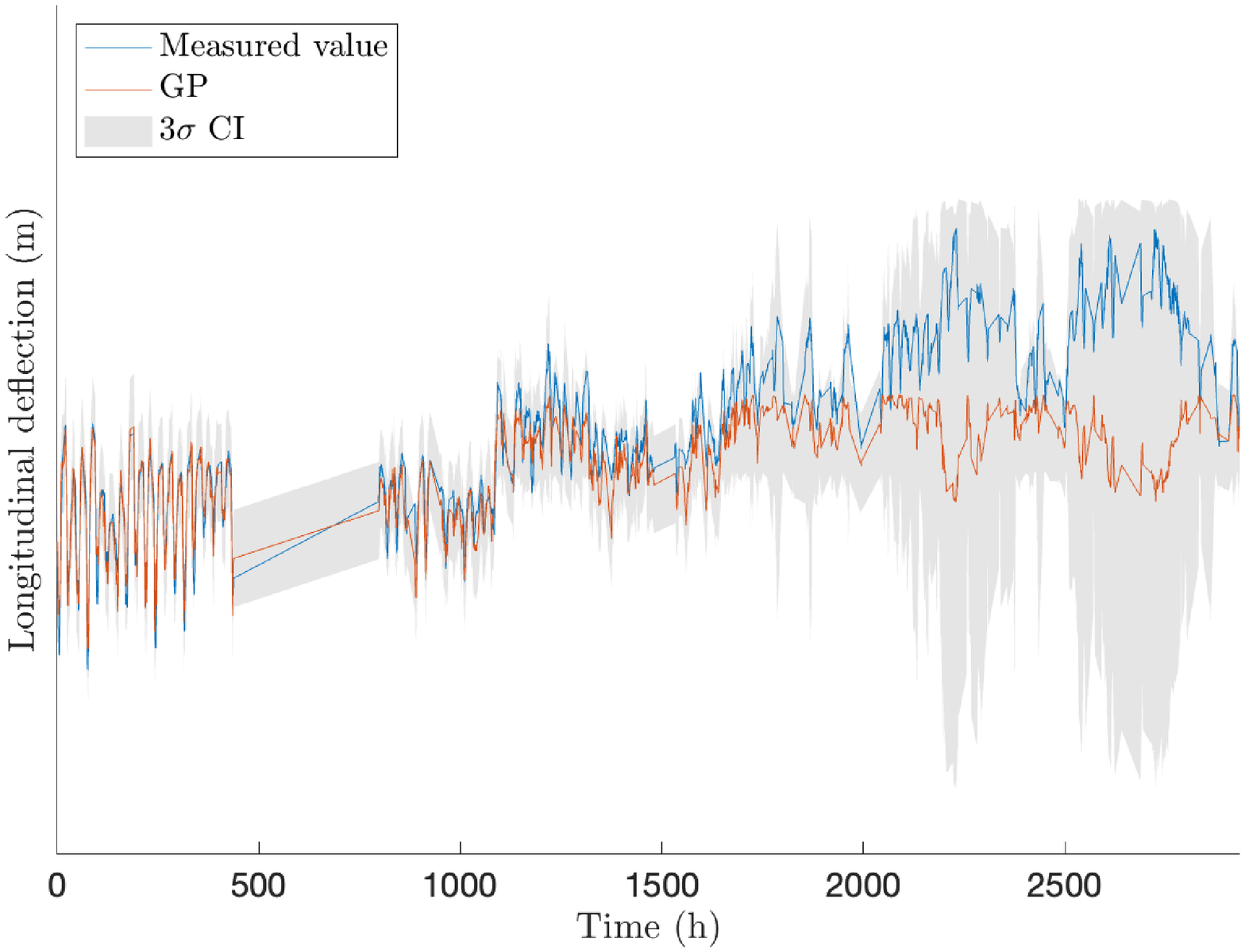}
\caption{}
\label{fig:tamar_black}
\end{subfigure}
\hspace{0.2cm}
\begin{subfigure}[b]{0.6\textwidth}
\centering
\includegraphics[width=1\linewidth]{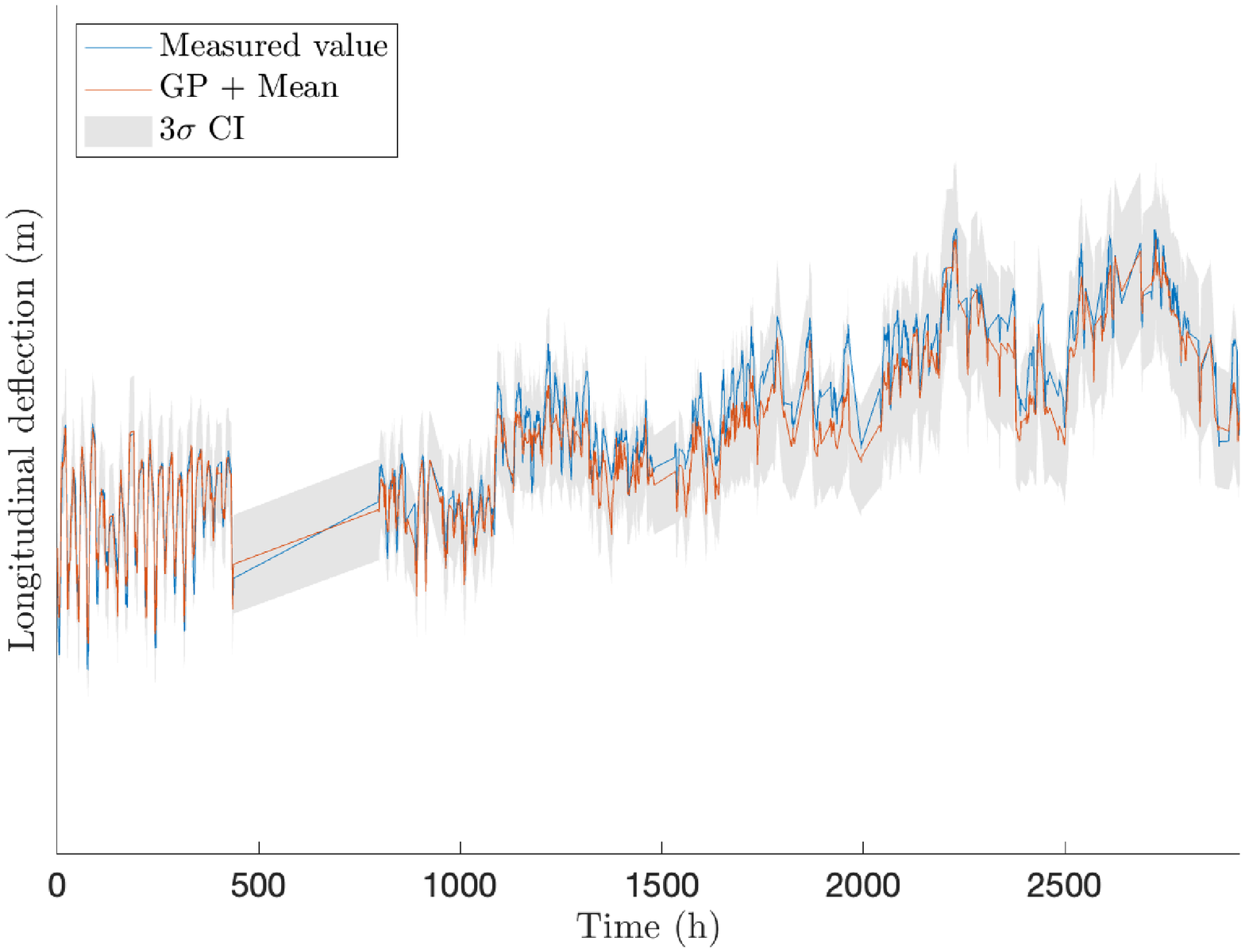}
\caption{}
\label{fig:tamar_grey}
\end{subfigure}
\caption{Model for bridge deck deflections; (a) shows the training and test datasets for the GPs, (b) is a GP predicion with a zero mean function prior and (c) shows the prediction when a  simple physics-informed mean function is incorporated. See \cite{zhang2020gaussian} for more details.}
\label{fig:tamar}
\end{figure}

The Tamar bridge is a cable-supported suspension bridge connecting Saltash 
and Plymouth in the South West of England which has been monitored by the 
Vibration Engineering Section at the University of Exeter 
\cite{koo2013structural}. The interest here is in the development of a model 
to predict bridge deck deflections that can be used as a performance 
indicator (see \cite{cross2012filtering,cross2012structural}). The variation 
in deck deflections are driven by a number of factors, including fluctuating 
temperature and loading from traffic (which are included as inputs to the  
model). Figure \ref{fig:tamar_data} shows the 
regression target considered in this example, which is a longitudinal 
deflection. The monitoring period shown is from September (Autumn) to January 
(Winter). In this figure one can see short term fluctuations (daily) and a 
longer term trend which is seasonal and driven by the increased hogging of 
the bridge deck as the ambient temperature decreases into the winter months. 
To mimic the situation where only a limited period of monitoring data is 
available for the establishment of an SHM algorithm, data from the initial 
month of the monitoring period is used to establish a GP regression model for 
deflection prediction (see \cite{zhang2020gaussian} for more details). 

A GP prediction, using the standard approach of a zero mean prior, is shown 
in \ref{fig:tamar_black}. Here one can see exactly the behaviour that is 
expected; the model is able to predict the deck deflections well in and 
around the training period, but is unable to predict the deflections in 
colder periods towards the end of the time series. The confidence intervals 
widen to reflect that the inputs to the model towards the end of the period 
are different from those in the training set - this demonstrates the 
usefulness of the GP approach, as one knows to place less trust in the 
predictions from 
this period.

To formulate a grey-box model for this scenario, a physics-informed prior 
mean function is adopted that encodes the expected linear expansion behaviour 
of stay-cables with temperature \cite{westgate2012environmental}. Figure 
\ref{fig:tamar_grey} shows the GP prediction with a linear prior mean 
function, where one can see a significant enhancement of predictive 
capability across the monitoring period. Where temperatures are at their 
lowest, the model predictions fall back on the prior mean function allowing 
some extrapolative capability. The prediction error is significantly smaller 
for the grey-box model in this case; the `black-box' nMSE is 68.65, whereas 
the GP with the physics-informed mean function has an nMSE of 7.33.

\subsubsection{Residual modelling for wave loading prediction}
In this example, we follow a similar approach of adopting a physics-informed 
mean function, this time with a dynamic Gaussian process formulation, a 
GP-NARX \cite{worden2018confidence}, to enhance predictive capability for a 
wave loading assessment. The monitoring or prediction of the loads a 
structure experiences in service is an important ingredient for health 
assessment, particularly where one wishes to infer e.g. fatigue damage 
accrued. 

The implementation of residual modelling is most effective where the assumptions and limitations of the white-box model are well understood. As a widely used method for wave loading prediction, Morison's Equation \cite{Morison1950} is employed here as a physics-informed mean function. This empirical law is known to simplify the behaviour of wave loading, not accounting for effects such as vortex shedding or other complex behaviours \cite{MorisonVortex} and will typically have residual errors in the region of \(20\%\)\cite{CoastalHydraulics}. Here we consider the addition of a data-based GP-NARX to a simplified version of Morison's Equation in an attempt to account for these missing phenomena. The model used is:
\begin{equation}
y_t =  \underbrace{C_d'U_t|U_t| + C_m'\dot{U}_t}_{Morison's\, Equation} + \;\: \underbrace{f([u_{t},\ u_{t-1}, ...,\ u_{t-l_u},\ y_{t-1},\ y_{t-2}, ...,\ y_{t-l_y}]) + \varepsilon}_{GP-NARX}
\end{equation}
where \(y_t\) is the wave force, \(C_d'\) is the drag coefficient, \(C_m'\) 
is the inertia coefficient, \(U\) is the wave velocity, \(\dot{U}\) is the 
wave acceleration, \(u_{t:t-l_u}\) are lagged exogeneous inputs and 
\(y_{t-1:t-l_u}\) are the lagged wave force, see \cite{pitchforth2021grey} 
for more details (this paper also shows an example of an input augmentation 
model, where Morison's equation 
is used as an additional input to the GP-NARX).

\begin{figure}[h]
\centering
\includegraphics[width=0.7\textwidth]{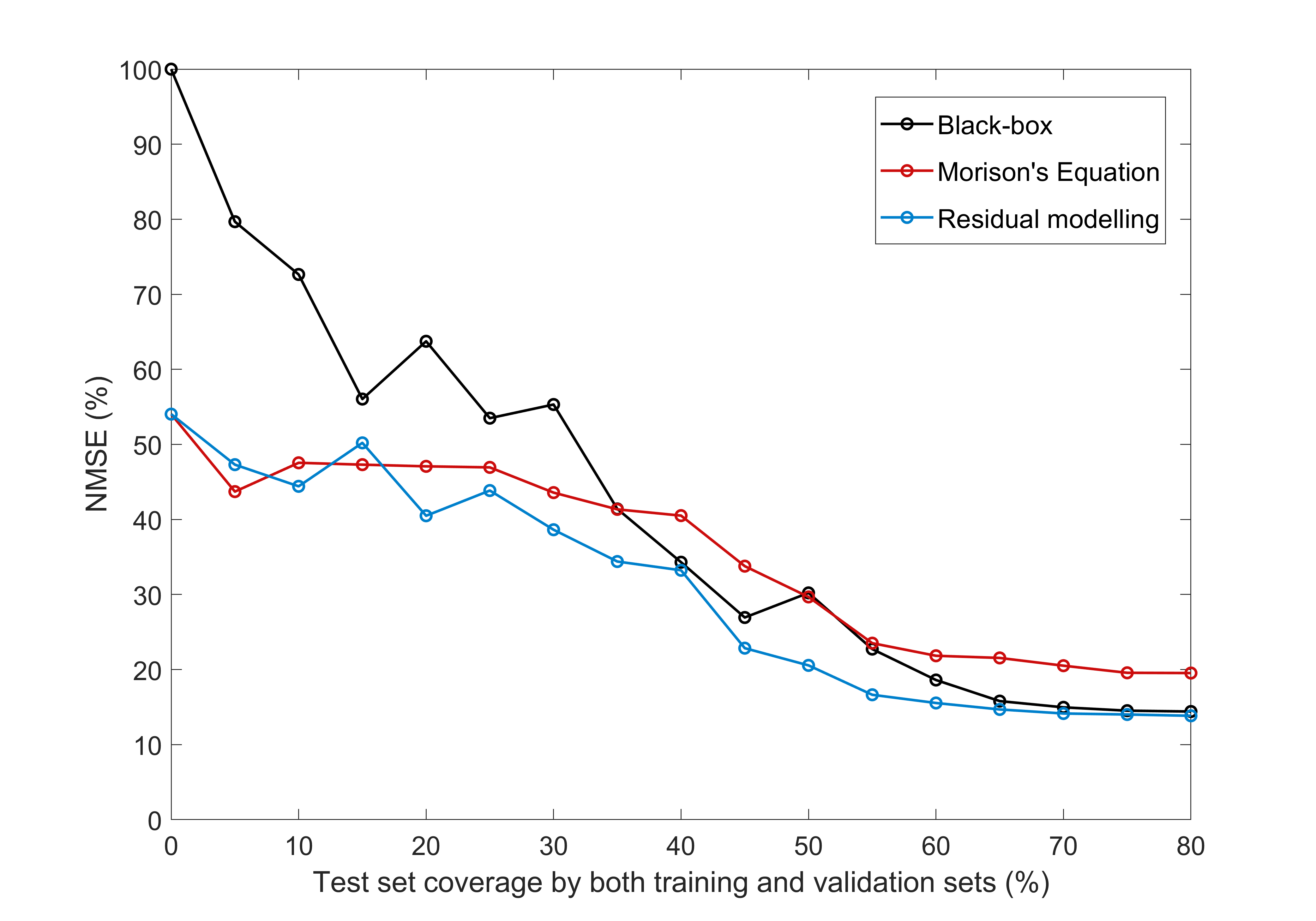}
\caption{A comparison of wave loading prediction model NMSEs vs test set coverage. Increasing coverage of the test set by the training and validation sets results in an increased level of model interpolation. See \cite{pitchforth2021grey} for more details.}
\label{fig:Residual_Black-box_Morison_NMSE_vs_Coverage}
\end{figure}

The Christchurch bay dataset is used here as an example to demonstrate the 
approach \cite{najafian2000uk}. To explore the generalisation capability with 
and without the physics-informed mean, different training sets for the 
GP-NARX are considered with increasing levels of coverage of the input space. 
A comparison of model errors (nMSE) with different training datasets is shown 
in Figure \ref{fig:Residual_Black-box_Morison_NMSE_vs_Coverage}. The coverage 
level is indicated as a percentage of the behaviour observed in the testing 
set that is also encountered in the training set \cite{pitchforth2021grey}. %
 
As in the Tamar Bridge example, the model structure offers a significant  
improvement in extrapolation, where testing conditions are different 
from those in training dataset. Following this approach allows predictions to 
be informed by the prior mean in the absence of evidence from data. Clearly 
the prior specification is very important in this case and a misspecified 
prior could do more harm than good. Once again we advocate the use of simple 
and well founded physics-based models in an attempt to avoid this issue. 

\subsection{Physics-derived covariance functions}

As discussed above, in a standard approach to GP regression, a generic covariance function such as a squared-exponential or one from the Mat\'{e}rn class is selected as a prior. In the posterior GP, the mean is a weighted sum of observations in the training set (see appendix), with the weightings provided by the covariance function and associated matrix. These commonly used functions encode that the covariance between points 
with similar inputs will be high and this allows the model to be 
data-driven in nature. 

In the case where one has some knowledge of a process of interest, it is possible to derive a covariance function that reflects this. As an example, in \cite{haywood2021structured}, a composite covariance function is designed to reflect the characteristics of the guided waves being modelled.

For some stochastic processes, the (auto)covariance can be directly derived from the equation of motion of a system. An example relevant for vibration-based SHM is the single degree of freedom (SDOF) oscillator
\begin{equation}
m\ddot{{y}}(t)+c\dot{{y}}(t)+k{y}(t)={F}(t) 
\label{eq:sdof}
\end{equation}
with mass, damping and stiffness parameters, $m,c,k$ respectively driven by a  
forcing process $F(t)$. In the case where the forcing is Gaussian white 
noise, the response $Y$ is a Gaussian process with (auto)covariance
\begin{equation}
\phi_{Y(\tau)}=\mathbb{E}[Y(t_1)Y(t_2)]=\frac{\sigma^2}{4m^2{\zeta\omega_n}^3}e^{-\zeta\omega_n|\tau|}
(\cos(\omega_d\tau)+\frac{\zeta\omega_n}{\omega_d}\sin(\omega_d|\tau|))
\label{eq:thecovariance}
\end{equation}
where standard notation has been used; $\omega_n=\sqrt{k/m}$, the natural 
frequency, $\zeta=c/2\sqrt{km}$, the damping ratio, 
$\omega_d=\omega_n\sqrt{1-\zeta^2}$, the damped natural frequency. See 
\cite{cross2021physics} and also 
\cite{papoulis2002probability,caughey1971nonlinear}. 

This covariance function can be readily used in the regression context and 
provides a useful prior process for oscillatory systems with a response 
dominated by a single frequency. This form of covariance function can be 
described as \textit{expressive} \cite{wilson2013gaussian} and proves useful 
even when the equation of motion of the system of interest differs from an 
SDOF linear assumption. 

\begin{figure}[h]
	\centering
	\includegraphics[width=0.7\textwidth]{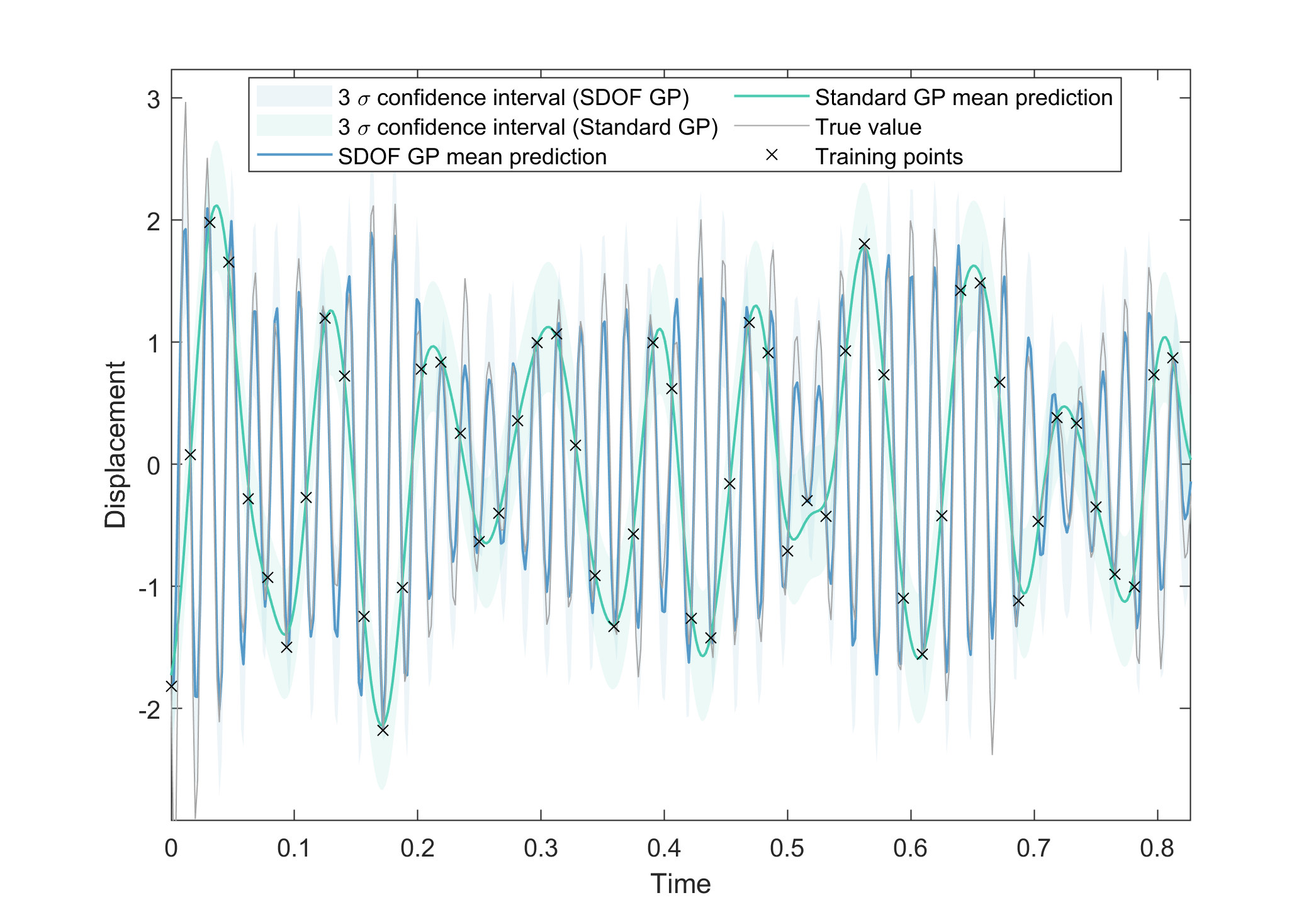}
	\caption{A comparison of a GP regression using derived and squared-exponential covariance function. See \cite{cross2021physics} for more details.}
	\label{fig:sdofcov}
\end{figure}

Figure \ref{fig:sdofcov} shows an example of a GP regression for a system 
with a cubic nonlinearity. Here the linear prior provides an appropriate 
structure for the regression and is flexible enough to incorporate the 
nonlinearity in the response. This is a simulated example with the GP 
training data shown with crosses in the figure (every 8th point)  - here the nMSE is 8.09. For comparison, a GP with a squared-exponential (SE) 
covariance function is established with the same training data. The SE 
process smooths through the data as expected (nMSE=66.7), whereas the derived covariance 
provides structure through the prior, resulting in good prediction 
during interpolation. The hyperparameters in the SDOF covariance function are 
physically interpretable, we are, therefore, able to guide their 
optimisation by providing the likely ranges for the system of interest. Here 
the benefit of being able to prescribe the likely frequency content of the 
system within the prior is clear and provides much advantage over the 
black-box approach - see \cite{cross2021physics} for more 
details.

\section{Constrained Gaussian processes}
\label{sec:constrain}

In scenarios where one lacks significant knowledge of the governing equations and solutions, grey-box methods that tend towards the black end of the spectrum can be particularly useful. An example of such approaches are constrained machine learners, which, very generally, aim to embed physical constraints into the learning procedure such that predictions made by the black-box model then adhere to these constraints. 

\begin{wrapfigure}{r}{0.39\textwidth}
	\centering
	\includegraphics[scale=0.45]{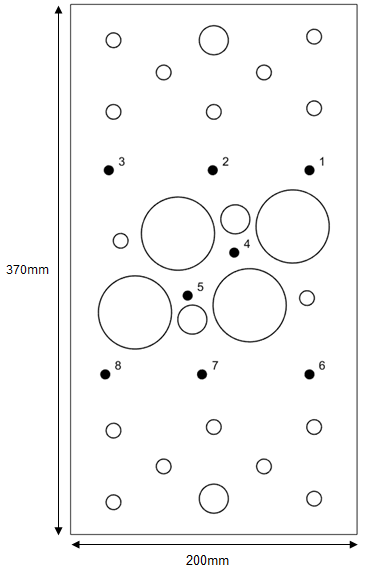}
	\caption{Test structure schematic with sensor locations, recreated from \cite{hensman2010locating}.}
	\label{fig:matt1}
\end{wrapfigure}
In the context of Gaussian process regression, there are a number of ways of constraining predictions, the simplest of which is to include data from boundaries within the model training. Other methods rely on applying constraints to the covariance function in a multiple output setting (see e.g. \cite{solin2018modeling, wahlstrom2013modeling,jidling2018probabilistic} and \cite{cross2019grey} where we employ derivative boundaries for beam deflection predictions).

Here we show an example of building known boundaries (geometry) into a GP regression via a sparse approximation of the covariance function. The approximation method relies on an eigendecomposition of the Laplace operator of a fixed domain \cite{solin2020hilbert}:
\begin{equation}\label{eq:matt1}
k(\mathbf{x},\mathbf{x}') \approx \sum_{i}^mS(\sqrt{\lambda_{i}})\phi_{i}(\mathbf{x})\phi_{i}(\mathbf{x}'),
\end{equation}
with $\phi_{i}$ and $\lambda_{i}$ the eigenfunctions and values, and $S$ the 
spectral density of the covariance function. If one chooses the fixed domain 
to reflect the geometry of the problem of interest, then inference with this 
model is appropriately bounded (see 
\cite{jones2020constraining,jones2021bayesian} for more details).

As an example here we employ constraints for a crack localisation problem via 
measurement of Acoustic Emission (AE). The localisation approach taken is to 
use artificial source excitations and an interpolating GP to provide a map of 
the differences in times of arrival ($\Delta$$T$) of AE sources to fixed 
sensor pairings across the surface of the structure 
\cite{hensman2010locating,jones2020bayesian}. Once constructed, the map can 
be used to assess the most likely location of any new AE sources. The bounded 
GP approximation allows one to build in the geometry of the structure under 
consideration.

To investigate the predictive capability of the constrained GP, a case study 
using a plate with a number of holes in is adopted. The holes, as shown in 
Figure \ref{fig:matt1}, provide complexity to the modelling challenge, 
introducing several complex phenomena such as wave mode conversion and signal 
reflection. Depending on the location of the source and sensor, the holes may 
also shield a direct propagation path to the receiver 
\cite{hensman2010locating}, adding further complication. 

\begin{figure}[H]
\centering
\includegraphics[scale=0.6]{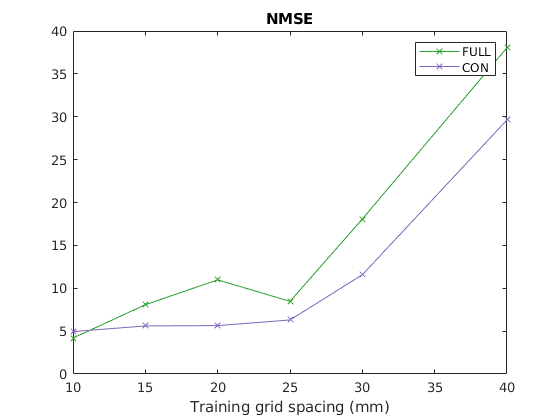}
\caption{Comparison between models errors for standard and bounded GPs for AE source localisation study. The nMSE is averaged across all sensor pair models for each training set considered.}
\label{fig:matt4}
\end{figure}

Neumann boundaries are imposed here around each hole and at the edge of the 
plate. To compare the performance of the standard and bounded GPs, differing 
amounts/coverage of artificial source excitations were used for model 
training. The initial characterisation of a structure via artificial source 
excitation can be expensive and time consuming, for structures in operation 
it may also be infeasible to access all areas/components. To mimic the 
scenario where it is not possible to collect artificial source excitations 
across a whole structure, here we restrict the training grid to excitation 
points in 
the middle of the plate. Figure \ref{fig:matt4} compares 
the performance of the standard and bounded GPs with training sets of varying 
grid densities. For each training set, the prediction error (nMSE) on the 
test set is averaged across every sensor pair (there are 8 sensors).

\begin{figure}[h]
\begin{center}
\includegraphics[width=0.9\textwidth]{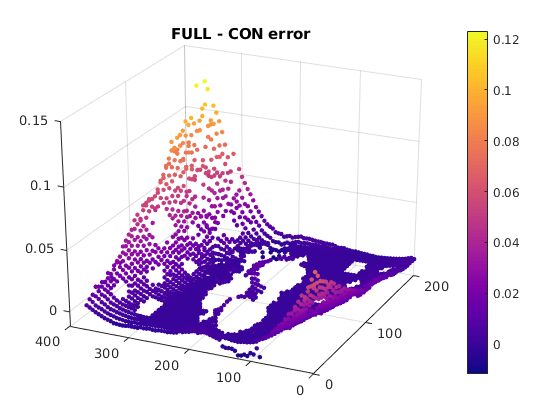}
\caption{Mapping of the difference between the full GP squared error and the constrained GP squared error across the test set for sensor pair 4-8.} \label{fig:matt3}
\end{center}
\end{figure}

From Figure \ref{fig:matt4} one can see that as the training set size 
reduces, the constrained GP consistently outperforms the standard full GP. 
This is particularly encouraging as the bounded GP remains a sparse 
approximation. As is consistent with our earlier observations, the inbuilt 
physical insight aids inference where training data are fewer. Figure 
\ref{fig:matt3} shows the difference in prediction error across the 
plate for the standard and bounded GPs for the 20mm spacing training case and 
a single 
sensor pairing. In this case, the squared error of the full GP is subtracted 
from the squared error of the constrained GP, i.e. positive values indicate a 
larger error in the full GP, whilst a negative value expresses a larger error 
in the constrained GP.

The figure highlights the locations on the plate where the constrained GP more accurately predicts the true $\Delta$$T$ values. As expected, the locations at which this effect is most prominent are those that move further away from the training points, and particularly towards the extremities of the domain. At these locations, it is clear that the additional physical insight provided by the constrained GP is able to enhance the predictive performance in comparison to the pure black-box model.  

\section{Gaussian processes in a state-space approach}
\label{sec:ss}

One of the canonical forms for dynamic models, in structural mechanical systems and beyond, is the state-space representation of the behaviour of interest. %

In the context of this work, the state space model (SSM) is considered to be a probabilistic object defined by two key probability densities; a transition density $\transdens$ and observation density $\obsdens$. %
The transition density relates the hidden states at a given time $\vec{x}_{t+1}$ to their previous possible values\footnote{The notation \emph{subscript} $a:b$ is used to denote values in that range inclusively, e.g.\ $\vec{x}_{0:t}$ is the value of the states $\vec{x}$ at all times from $t=0$ to $t=t$.}, $\vec{x}_{0:t}$ and previous external inputs to the system, e.g.\ forcing, $\vec{u}_{0:t}$. The observation model relates available measurements $\vec{y}_{t}$ to the hidden states $\vec{x}_{t}$, which may also be dependent on the external inputs at that time $\vec{u}_t$.

The state space formulation can be used to properly account for 
measurement noise (filtering and smoothing), and is commonly used for 
parameter estimation (the well-known Kalman filter is a closed form solution 
for linear and Gaussian systems). Use of the state-space models as a grey-box 
formulation in this setting is common within the control community
\cite{kristensen2004parameter,tulleken1993grey}.

Here, we are interested in the case where we only have partial knowledge of a 
system - this could take the form of missing or incorrect physics in the 
equations of motion, or could be a lack of access to key measurements such as 
the force a system undergoes. In Section \ref{sec:bayes} we considered a 
GP-NARX 
formulation for wave loading prediction, with the ultimate aim of informing a 
fatigue 
assessment. The state-space formulation shown here offers an alternative 
means for load 
estimation which simultaneously provides parameter and state estimation in 
a Bayesian setting.

Joint input-state and input-state-parameter problems have seen growing 
interest in recent years, see, for example 
\cite{lourens2012joint,azam2015dual,naets2015online,maes2019tracking,dertimanis2019input}.
The approach shown here is one that considers a representation of a Gaussian 
process within a state space formulation to model the unknown forcing 
(following \cite{alvarez2009latent,hartikainen2012sequential}). This is 
achieved by deriving the transfer function of a Mat\'{e}rn kernel (via its 
spectral density), which provides a flexible model component to account for 
the unmeasured behaviour. Inference over the state space model is via Markov 
Chain Monte Carlo to provide distributions for parameter and hyperparameter 
estimations.

\begin{figure}[h]
\centering
\includegraphics[width=0.9\textwidth]{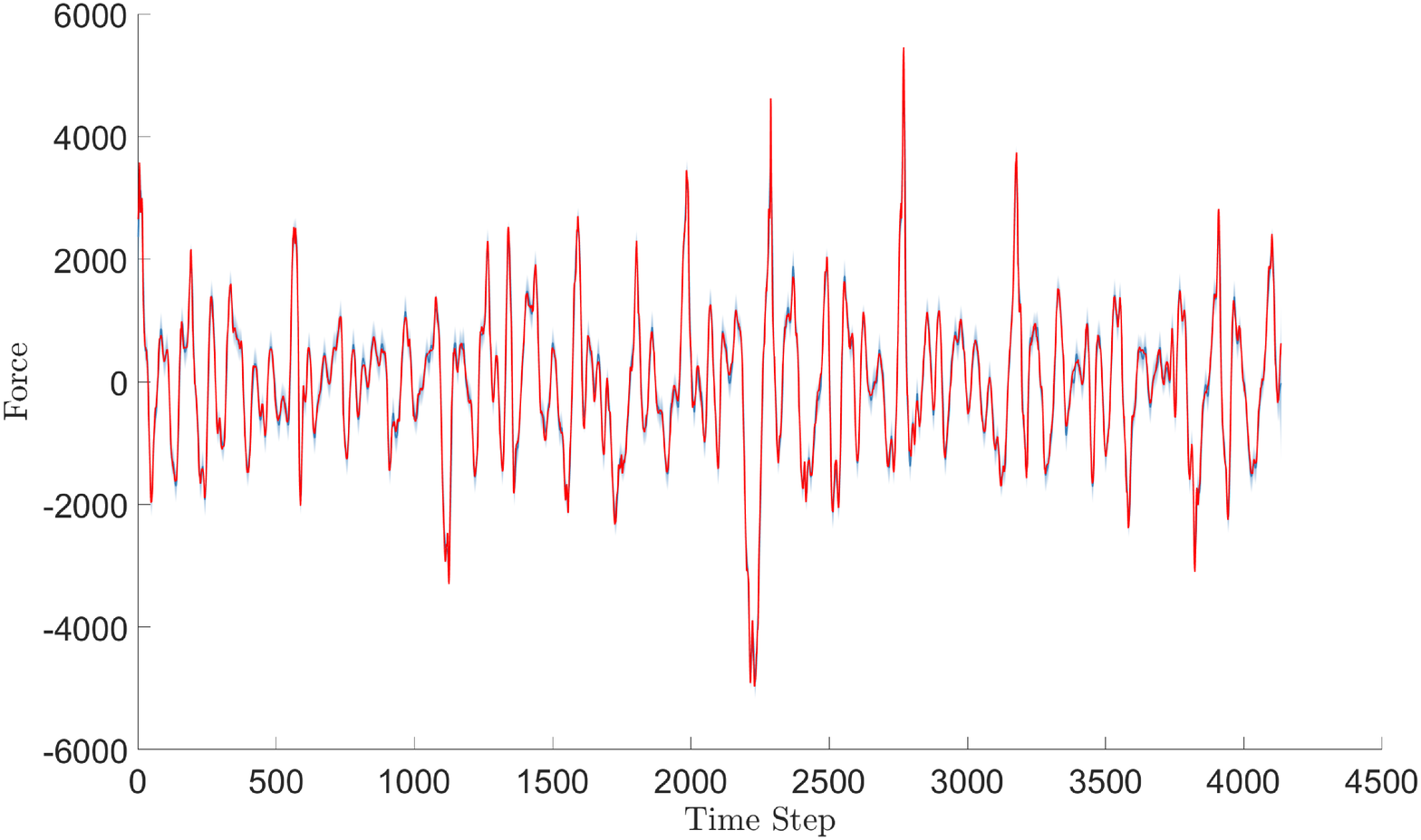}
\caption{Force estimation for MDOF simulation under Christchurch bay 
loading time history\cite{rogers2020application}}
\label{fig:SSMforce}
\end{figure}

Figure \ref{fig:SSMforce} shows an example of force recovery for a simulated 
multi-degree of 
freedom system excited by a forcing time history from the Christchurch bay 
example discussed in Section \ref{sec:bayes}. Here one can see that the force 
has been accurately inferred, the nMSE in this case is 1.15. For more details 
and analysis see \cite{rogers2020application}.

The inference problem becomes significantly more challenging for nonlinear 
systems, and even more so if our knowledge of that nonlinearity is 
incomplete. Recently \cite{rogers2020bayesian} has attempted this extension 
for the input-state estimation case for a known nonlinearity. The difficulty 
in inference is met there by employing a methodology based on Sequential 
Monte Carlo, specifically Particle Gibbs with Ancestor Sampling, to allow 
recovery of the states and the hyperparameters of the GP.

In the face of an unknown nonlinearity this framework may also be employed, 
where the GP may be used to account for missing behaviours from the assumed 
equations of motion. In \cite{friis2020gaussian}, it is shown how this 
approach can be applied to a Duffing oscillator to learn the unknown cubic 
component of the model in a Bayesian manner without requiring prior knowledge 
of the nonlinear function. %
There are two particular advantages to this approach, the first is that it 
allows nonlinear system identification with a linear model, the Kalman filter 
and RTS smoother, which has significant computational advantages and arguably 
makes ``fully Bayesian'' inference industrially feasible. %
The second benefit is that it allows a user to apply very weak prior 
knowledge about the nonlinearity; in the language of grey-box models, there 
is a strong white box component (the second order linear system) but the form 
of the nonlinearity is the very flexible nonparametric GP. %
Contrast this approach with the purely black-box alternative of the GP-SSM, see \cite{frigola2015bayesian}, which suffers from significant nonindentifiability and computational challenges.

\section{Conclusions}

This chapter has introduced and demonstrated physics-informed machine learning methods suitable for SHM problems and inference in structural dynamics more generally. The methods allow the embedding of one's physical insight of a structure or system into a data-driven assessment. The resulting models have proven to be particularly useful in situations where  training data are not available across the operational envelope - a common occurrence in structural monitoring campaigns. 

The Bayesian approach adopted in Section \ref{sec:bayes} allows predictions 
to fall back on a prior physical model in the absence of evidence from data. 
This pragmatic approach proved useful in the examples shown here but does 
rely on trusting the physical model in extrapolation. The ability to 
constrain the Gaussian process prior to known boundary conditions shown in 
Section \ref{sec:constrain} requires less physical insight and gives both an 
improved modelling performance, as well as providing the guarantee that 
predictions made adhere to known underlying physical laws of the system under 
consideration. At the whiter end of the spectrum, the state-space examples 
discussed in Section \ref{sec:ss} provide a principled means of inference 
over structures with unknown forcing or nonlinearites. Some examples of where 
the presented methodology may be of benefit could include better 
understanding of fatigue damage accrual and parameter identification for e.g. 
novelty/damage detection.

As well as providing an enhanced predictive capability, the models introduced 
here have the benefit of being more readily interpretable than their purely 
black-box counterparts. In the past, a barrier to the uptake of SHM 
technology has been the lack of trust owners and operators have in so-called 
black-box models. Perhaps naturally, there is a hesitancy to adopt algorithms 
not derived from physics-based models, but this may also be due, in part, to 
their misuse in the past. We hope that this will be ameliorated by more 
interpretable models which also have the benefit of being more easily 
optimised (Section \ref{sec:bayes} shows an example where the hyperparameters 
in the GP regression take on physical meaning).

Physics-informed machine learning is rapidly becoming a popular research 
field in its own right, with many promising results and avenues for 
investigation. This review paper \cite{willard2020integrating} currently on 
arXiv has 300 references largely populated by papers from the last two years. 
It is likely that many of the emerging methods will prove useful in SHM. The 
work here has focussed on a Gaussian process framework, clearly the use of 
neural networks provide an alternative grey-box route, as these are also 
commonly used in our field. We look forward to seeing how these may be 
adopted for SHM tasks.

\subsubsection{Acknowledgements} We would like to thank Keith Worden for his 
general support and also particularly for provision of the Christchurch Bay 
data and support of the state-space work. Additionally, thanks is offered to 
James Hensman, Mark Eaton, Robin Mills, Gareth Pierce and Keith Worden for 
their work in acquiring the AE data set used here.  We would like to thank 
Ki-Young Koo and James Brownjohn in the Vibration Engineering Section at the 
University of Exeter for provision of the data from the Tamar Bridge. Thanks 
also to Steve Reed, formerly of DSTL for the provision of the Tucano data.
Finally, the 
authors would like to acknowledge the support of the EPSRC, particularly 
through grant reference number EP/S001565/1, and Ramboll Energy for their 
support of SG and DP.

\appendix
\section{Gaussian Process Regression}
Here we follow the notation used in \cite{Rasmussen2006}; 
$k({\bf{x}}_p,{\bf{x}}_q)$ defines a covariance matrix $K_{pq}$, with 
elements evaluated at the points ${\bf x}_p$ and ${\bf x}_q$, where 
$\bf{x_i}$ may be multivariate. 

Assuming a zero-mean function, the joint Gaussian distribution between 
measurements/observations $\bf{y}$ with inputs $X$ and unknown/testing 
targets ${\bf y}^*$ with inputs $X^*$ is 
\begin{equation}
\begin{bmatrix}
{\bf y} \\
{\bf y}^*
\end{bmatrix}
\sim \mathcal{N} \left( 0,
\begin{bmatrix}
K(X,X)+\sigma_n^2I &  K(X,X^*) \\
K(X^*,X) & K(X^*,X^*) 
\end{bmatrix}
\right)
\label{eq:Sjoint}
\end{equation}
The distribution of the testing targets ${\bf y}^*$ conditioned on the 
training data (which is what we use for prediction) is also Gaussian:
\begin{equation}
\begin{aligned}
{\bf y}^*|&X_*,X,{\bf y} \sim \mathcal{N}(K(X^*,X)(K(X,X)+\sigma_n^2I)^{-1} 
{\bf y}, \\
&K(X^*,X^* )-K(X^*,X)(K(X,X)+\sigma_n^2I)^{-1} K(X,X^*))
\end{aligned}
\label{eq:condcvfn}
\end{equation}
See \cite{Rasmussen2006} for the derivation. The mean and covariance here are 
that of the posterior Gaussian process. In this work covariance function 
hyperparameters are sought by 
maximising the marginal likelihood of the predictions
\begin{equation}
\log p({\bf y}|X, \boldsymbol{\theta})=-\frac{1}{2} {\bf y}^T K^{-1} {\bf 
y} - \frac{1}{2} \log |K| - \frac{n}{2} \log 2\pi
\end{equation}
via a particle swarm optimisation\footnote{Yes, we could be more Bayes here} 
\cite{rogers2019towards}.

\bibliographystyle{unsrtnat}
\bibliography{book_chapter}
\end{document}